\newcounter{licntr}			

\newcommand{\stepli}{\refstepcounter{licntr}}	

\newcommand{\lival}{\thelicntr}			
\newenvironment{liout}{
		\stepli			
		\samepage			
		\begin{list}{		
			(\lival)\hfill}{} 	
			\samepage
			\item 			
		}{ \end{list}}

\newenvironment{li}{
		\begin{liout} 			
		\begin{list}{\samepage \noindent}{} 	
		\item}{				
		\end{list} 
		\end{liout}}

\newenvironment{li*}{
	\stepli
	\samepage				
	\begin{trivlist} \item[] (\lival) \end{trivlist}
	\begin{trivlist} 
	{\samepage \item[]}
	}{\end{trivlist}}					

\newcounter{alphcntr}		

\documentclass[11pt,a4paper]{article}
\usepackage[hyperref]{emnlp2018}
\usepackage{times}
\usepackage{latexsym}
\usepackage{graphicx}
\usepackage{url}
\usepackage[small,bf]{caption} 

\usepackage{url}

\aclfinalcopy 

\title{Getting Gender Right in Neural Machine Translation}

\author{Eva Vanmassenhove$^\alpha$\\
  \And
  Christian Hardmeier$^\beta$\\
  \\
   $^\alpha$ ADAPT, School of Computing,
  Dublin City University, Dublin, Ireland\\
  {\tt firstname.lastname@adaptcentre.ie} \\
  \\
   $^\beta$ Department of Linguistics and Philology, 
  Uppsala University, Uppsala, Sweden\\
  {\tt christian.hardmeier@lingfil.uu.se} \And
  Andy Way$^\alpha$\\
  }
\date{}

\begin{document}
\maketitle
\begin{abstract}

Speakers of different languages must attend to and encode strikingly different aspects of the world in order to use their language correctly ~\cite{Sapir1921,Slobin1996}. One such difference is related to the way gender is expressed in a language. Saying ``I am happy'' in English, does not encode any additional knowledge of the speaker that uttered the sentence. However, many other languages do have grammatical gender systems and so such knowledge would be encoded. In order to correctly translate such a sentence into, say, French, the inherent gender information needs to be retained/recovered. The same sentence would become either ``Je suis heureux'', for a male speaker or ``Je suis heureuse'' for a female one. Apart from morphological agreement, demographic factors (gender, age, etc.) also influence our use of language in terms of word choices or even on the level of syntactic constructions~\cite{Tannen1991,Pennebaker2003}.
We integrate gender information into NMT systems. Our contribution is two-fold: (1) the compilation of large datasets with speaker information for 20 language pairs, and (2) a simple set of experiments that incorporate gender information into NMT for multiple language pairs. Our experiments show that adding a gender feature to an NMT system significantly improves the translation quality for some language pairs. 
\end{abstract}

\section{Introduction}
In the field of linguistics, the differences between male and female traits within spoken and written language have been studied both empirically and theoretically, revealing that the language used by males and females differs in terms of style and syntax~\cite{Coates2015}. The increasing amount of work on automatic author classification (or `author profiling') reaching relatively high accuracies on domain-specific data corroborates these findings~\cite{Rangel2013,santosh2013}. However, determining the gender of an author based solely on text is not a solved issue. Likewise, the selection of the most informative features for gender classification remains a difficult task~\cite{Litvinova2016}.

When translating from one language into another, original author traits are partially lost, both in human and machine translations~\cite{Mirkin2015,Rabinovich2016}. However, in the field of Machine Translation (MT) one of the most observable consequences of this missing information are morphologically incorrect variants due to a lack of agreement in number and gender with the subject. Such errors harm the overall fluency and adequacy of the translated sentence. Furthermore, gender-related errors are not just harming the quality of the translation as getting the gender right is also a matter of basic politeness. Current systems have a tendency to perpetuate a male bias which amounts to negative discrimination against half the population and this has been picked up by the media.\footnote{\url{https://www.theguardian.com/technology/2017/apr/13/ai-programs-exhibit} \url{-racist-and-sexist-biases-research}} 

Human translators rely on contextual information to infer the gender of the speaker in order to make the correct morphological agreement. However, most current MT systems do not; they simply exploit statistical dependencies on the sentence level that have been learned from large amounts of parallel data. Furthermore, sentences are translated in isolation. As a consequence, pieces of information necessary to determine the gender of the speakers, might get lost. The MT system will, in such cases, opt for the statistically most likely variant, which depending on the training data, will be either the male or the female form. Additionally, in the field of MT, training data often consists of both original and translated parallel texts: large parts of the texts have already been translated, which, as studied by Mirkin et al.~(\citeyear{Mirkin2015}), does not preserve the original demographic and psychometric traits of the author, making it very hard for a Neural MT (NMT) system to  determine the gender of the author.

With this in mind, a first step towards the preservation of author traits would be their integration into an NMT system. As `gender' manifests itself not only in the agreement with other words in a sentence, but also in the choice of context-based words or on the level of syntactic constructions, the sets of experiments conducted in this paper focus on the integration of a gender feature into NMT for multiple language pairs. 

 The structure of the paper is the following: related work is described in Section~\ref{sec:relwork}; Section~\ref{sec:data} describes and analyses the datasets that were compiled; the experimental setup is discussed in Section~\ref{sec:exp}; the results are presented in Section~\ref{sec:results}; finally, we conclude and provide some ideas for future work in Section~\ref{sec:conclusions}.

\section{Related Work}\label{sec:relwork}

Differences in the language between male and female speakers have been studied within various fields related to linguistics, including Natural Language Processing (NLP) for author profiling, conversational agents, recommendation systems etc. Mirkin et al.~(\citeyear{Mirkin2015}) motivated the need for more personalized MT. Their experiments show that MT is detrimental to the automatic recognition of linguistic signals of traits of the original author/speaker. Their work suggests using domain-adaptation techniques to make MT more personalized but does not include any actual experiments on the inclusion of author traits in MT.

Rabinovich et al.~(\citeyear{Rabinovich2016}) conducted a series of experiments on preserving original author traits, focusing particularly on gender. As suggested by Mirkin et al.~(\citeyear{Mirkin2015}), they treat the personalization of Statistical MT (SMT) systems as a domain-adaptation task treating the female and male gender as two different domains. They applied two common simple domain-adaptation techniques in order to create personalized SMT: (1) using gender-specific phrase-tables and language models, and (2) using a gender-specific tuning set. Although their models did not improve over the baseline, their work provides a detailed analysis of gender traits in human and machine translation. 

Our work is, to the best of our knowledge, the first to attempt building a speaker-informed NMT system. Our approach is similar to the work of Sennrich et al.~(\citeyear{Sennrich2016control}) on controlling politeness, where some sentence of the training data are followed with an `informal' or `polite' tag indicating the level of politeness expressed.

\section{Compilation of Datasets}\label{sec:data}
One of the main obstacles for more personalized MT systems is finding large enough annotated parallel datasets with speaker information. Rabinovich et al.~(\citeyear{Rabinovich2016}) published an annotated parallel dataset for EN--FR and EN--DE. However, for many other language pairs no sufficiently large annotated datasets are available.

To address the aforementioned problem, we published online a collection of parallel corpora licensed under the Creative Commons Attribution 4.0 International License for 20 language pairs~\cite{Vanmassenhove2018}.\footnote{\url{https://github.com/evavnmssnhv/Europarl-Speaker-Information}} 

We followed the approach described by Rabinovich et al.~(\citeyear{Rabinovich2016}) and tagged parallel sentences from Europarl~\cite{Koehn2005} with speaker information (name, gender, age, date of birth, euroID and date of the session) by retrieving speaker information provided by tags in the Europarl source files. The Europarl source files contain information about the speaker on the paragraph level and the filenames contain the data of the session. By retrieving the names of the speakers together with meta-information on the members of the European Parliament (MEPs) released by Rabinovich et al.~(\citeyear{Rabinovich2016}) (which includes among others name, country, date of birth and gender predictions per MEP), we were able to retrieve demographic annotations (gender, age, etc.). An overview of the language pairs as well as the amount of annotated parallel sentences per language pair is given in Table~\ref{tbl:numberSents}.

\begin{table}[h!]
\centering
\footnotesize\begin{tabular}{|ll|ll|}
\hline
\bf{Languages}  &   \bf{\# sents}    &  \bf{Languages}  &   \bf{\# sents}   \\ \hline
\bf{EN--BG}	&	306,380	    &   \bf{EN--IT}	&	1,297,635	\\ \hline
\bf{EN--CS}	&	491,848	    &	\bf{EN--LT}	&	481,570	    \\ \hline
\bf{EN--DA}	&	1,421,197	&	\bf{EN--LV}	&	487,287	    \\ \hline
\bf{EN--DE} &	1,296,843	&	\bf{EN--NL}	&	1,419,359	\\ \hline
\bf{EN--EL}	&	921,540	    &	\bf{EN--PL}	&	478,008	    \\ \hline
\bf{EN--ES}	&	1,419,507	&	\bf{EN--PT}	&	1,426,043	\\ \hline
\bf{EN--ET}	&	494,645	    &	\bf{EN--RO}	&	303,396	    \\ \hline
\bf{EN--FI} &	1,393,572	&	\bf{EN--SK}	&	488,351	    \\ \hline
\bf{EN--FR}	&	1,440,620	&	\bf{EN--SL}	&	479,313	    \\ \hline
\bf{EN--HU}	&	251,833	    &	\bf{EN--SV}	&	1,349,472	\\ \hline

\end{tabular}
\caption{Overview of annotated parallel sentences per language pair}\label{tbl:numberSents}
\end{table}

\subsection{Analysis of the EN--FR Annotated Dataset}

We first analysed the distribution of male and female sentence in our data. In the 10 different datasets we experimented with, the percentage of sentences uttered by female speakers is very similar, ranging between 32\% and 33\%. This similarity can be explained by the fact that Europarl is largely a multilingual corpus with a big overlap between the different language pairs.

We conducted a more focused analysis on one of the subcorpora (EN--FR) with respect to the percentage of sentences uttered by males/females for various age groups to obtain a better grasp of what kind of data we are using for training. As can be seen from Figure~\ref{fig:analysisFM}, with the exception of the youngest age group (20--30), which represents only a very small percentage of the total amount of sentences (0.71\%), more male data is available in all age groups. Furthermore, when looking at the entire dataset, 67.39\% of the sentences are produced by male speakers. Moreover, almost half of the total number of sentences are uttered by the 50--60 age group (43.76\%).


{\footnotesize\begin{figure}[h]
\includegraphics[scale=0.70]{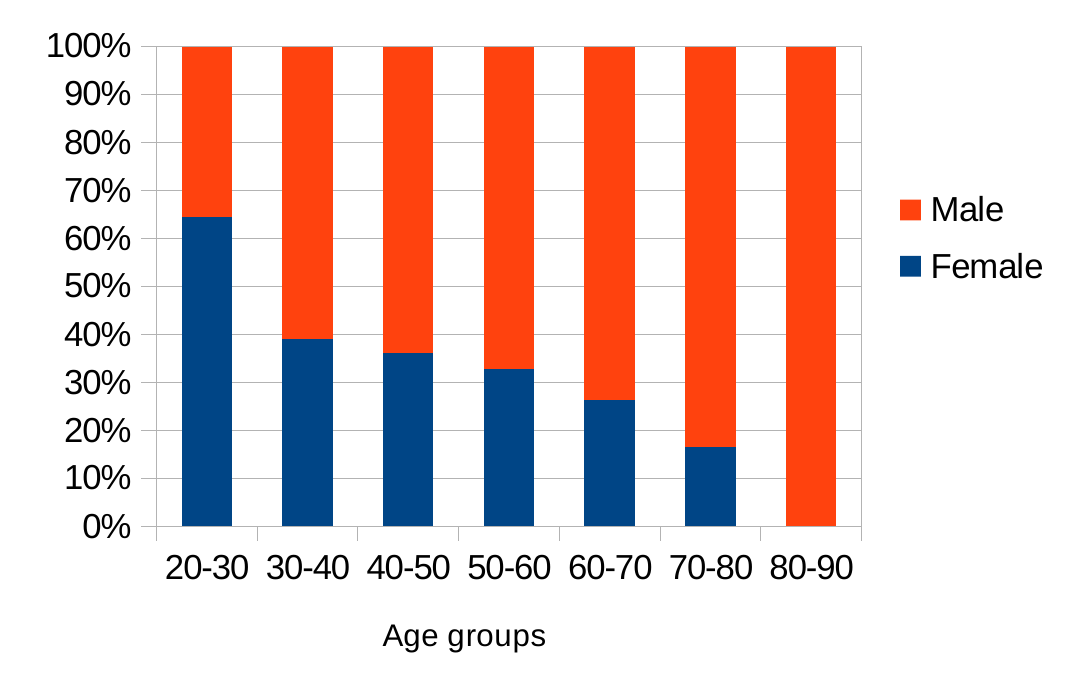}
\caption{Percentage of female and male speakers per age group}\label{fig:analysisFM}
\end{figure}}

The analysis shows that indeed, there is a gender unbalance in the Europarl dataset, which will be reflected in the translations that MT systems trained on this data produce.  

\section{Experimental Setup}\label{sec:exp}
\subsection{Datasets}
We carried out a set of experiments on 10 language pairs (the ones for which we compiled more than 500k annotated Europarl parallel sentences):  EN--DE, EN--FR, EN--ES, EN--EL, EN--PT, EN--FI, EN--IT, EN--SV, EN--NL and EN--DA. We augmented every sentence with a tag on the English source side, identifying the gender of the speaker, as illustrated in~(\ref{ExampleComplex}). This approach for encoding sentence-specific information for NMT has been successfully exploited to tackle other types of issues, multilingual NMT systems (e.g., Zero Shot Translation~\cite{GoogleZST}), domain adaptation~\cite{Sennrich2016control}, etc.
\begin{li}
{\small \centering
\begin{tabular}{l}
``FEMALE Madam President, as a...''
\label{ExampleComplex}
\end{tabular}}
\end{li}

For each of these language pairs we trained two NMT systems: a baseline and a tagged one. We evaluated the performance of all our systems on a randomly selected 2K general test set. Moreover, we further evaluated the EN--FR systems on 2K male-only and female-only test sets to have a look at the system performance with respect to gender-related issues. We also looked at two additional male and female test sets in which the first person singular pronoun appeared. 

\subsection{Description of the NMT Systems}
We used the OpenNMT-py toolkit~\cite{Klein2017} to train the NMT models. The models are sequence-to-sequence encoder-decoders with LSTMs as the recurrent unit~\cite{Bahdanau2014,Cho2014,Sutskever2014} trained with the default parameters. In order to by-pass the OOV problem and reduce the number of dictionary entries, we use word-segmentation with BPE ~\cite{Sennrich2015}. We ran the BPE algorithm with 89,500 operations~\cite{Sennrich2015}. All systems are trained for 13 epochs and the best model is selected for evaluation.

\section{Results}\label{sec:results}
In this section we discuss some of the results obtained. We hypothesized that the male/female tags would be particularly helpful for French, Portuguese, Italian, Spanish and Greek, where adjectives and even verb forms can be marked by the gender of the speaker. Since, according to the literature, women and men also make use of different syntactic constructions and make different word choices, we also tested the approach on other languages that do not have morphological agreement with the gender of the speaker such as Danish (DA), Dutch (NL), Finnish (FI), German (DE) and Swedish (SV). 

First, we wanted to see how our tagged systems performed on the general test set compared to the baseline. In Table~\ref{tbl:BLEU}, the BLEU scores for 10 baseline and 10 gender-enhanced NMT systems are presented.

\begin{table}[h!]
\centering
\footnotesize\begin{tabular}{|l|c|c|}
\hline
\bf{Systems}	&	\bf{EN}		&	\bf{EN-TAG}	            \\	\hline
\bf{FR} 		&	37.82	    &   \bf{39.26\rlap{*}}      \\	\hline
\bf{ES}	    	&	42.47		&	42.28 	 	            \\	\hline
\bf{EL}		    &	31.38		&	\bf{31.54}	 	        \\	\hline
\bf{IT}		    &	31.46		&	\bf{31.75\rlap{*}}	 	\\	\hline
\bf{PT}		    &	36.11		&	\bf{36.33}	 	        \\	\hline\hline
\bf{DA}		    &	36.69		&	\bf{37.00\rlap{*}}      \\	\hline
\bf{DE}		    &	28.28		&	28.05 	 	            \\	\hline
\bf{FI}	    	&	21.82		&	21.35\rlap{*}	 	    \\	\hline
\bf{SV}		    &	35.42		&	35.19 	 	            \\	\hline
\bf{NL}	    	&	28.35		&	28.22                   \\	\hline
\end{tabular}
 
 \caption{BLEU scores for the 10 baseline (denoted with \textbf{EN}) and the 10 gender-enhanced NMT (denoted with \textbf{EN-TAG}) systems. Entries labeled with * present statistically significant differences (p $<$ 0.05). Statistical significance was computed with the MultEval tool~\cite{Clark2011}. }\label{tbl:BLEU}
\end{table}

While most of the BLEU-scores~\cite{Papineni2002} in Table~\ref{tbl:BLEU} are consistent with our hypothesis, showing (significant) improvements for the NMT systems enriched with a gender tag (EN-TAG) over the baseline systems (EN) for French, Italian, Portuguese and Greek, the Spanish enriched system surprisingly does not (--0.19 BLEU). As hypothesized, the Dutch, German, Finnish and Swedish systems do not improve. However, the Danish (EN--DA) enriched NMT system does achieve a significant +0.31 BLEU improvement. 

We expected to see the strongest improvements in sentences uttered by female speakers as, according to our initial analysis, the male data was over-represented in the training. To test this hypothesis, we evaluated all systems on a male-only and female-only test set. Furthermore, we also experimented on test sets containing the pronoun of the first person singular as this form is used when a speaker refers to himself/herself. The results on the specific test set for the EN--FR dataset are presented in Table~\ref{tbl:testsets}. As hypothesized, the biggest BLEU score improvement is observed on the female test set, particularly, the test sets containing first person singular pronouns (F1).

\begin{table}[h!]
\centering
\footnotesize\begin{tabular}{|l|c|c|}
\hline
\bf{Test Sets}	    &	\bf{EN}		&	\bf{EN-TAG}	        \\	\hline
\bf{FR (M)}		    &	37.58		&	\bf{38.71*}	 	    \\	\hline
\bf{FR (F)} 		&	37.75	    &	\bf{38.97*}  	    \\	\hline
\bf{FR (M1)}	    &	39.00	    &	\bf{39.66*}	 	    \\	\hline
\bf{FR (F1)}		&	37.32		&	\bf{38.57*}   	 	\\ \hline
\end{tabular}
\caption{BLEU-scores on EN--FR comparing the baseline (EN) and the tagged systems (EN--TAG) on 4 different test sets: a test set containing only male data (M), only female data (F), 1st person male data (M1) and first person female data (F1). All the improvements of the EN-TAG system are statistically significant (p $<$ 0.5), as indicated by *.}\label{tbl:testsets}
\end{table}

We had a closer look at some of the translations.\footnote{We used the tool provided by Tilde \url{https://www.letsmt.eu/Bleu.aspx} to see where the BLEU score between the baseline and our tagged systems varied the most.} There are cases where the gender-informed (TAG) system improves over the baseline (BASE) due to better agreement. Interestingly, in~(\ref{exmpl:pres}) the French female form of vice-president (vice-pr\'{e}sidente) appears in the translation produced by the BASE system while the male form is the correct one. The gender-informed system does make the correct agreement by using the female variant. In~(\ref{exmpl:heureux}) the speaker is female but the baseline system outputs a male form of the adjective `happy' (`heureux'). 

\begin{li}
{\small \centering
\begin{tabular}{ll}
(Ref) & En tant que \emph{vice-pr\'{e}sident}...\\
(BASE) & En tant que \emph{vice-pr\'{e}sidente}...\\
(TAG) & En tant que \emph{vice-pr\'{e}sident}...\\
\end{tabular}}\label{exmpl:pres}
\end{li}

\begin{li}
{\small \centering
\begin{tabular}{ll}
(Ref) & ... je suis \emph{heureuse} que...\\
(BASE) & ... je suis \emph{heureux} que...\\
(TAG) & ... je suis \emph{heureuse} que...\\
\end{tabular}}\label{exmpl:heureux}
\end{li}

However, we also encountered cases where the gender-informed system fails to produce the correct agreement, as in~(\ref{exmpl:not}), where both the BASE and the TAG system produce a male form (`embarass\'e') instead of the correct female one (`embarass\'ee' or `g\^{e}n\'{e}e'). 

\begin{li}
{\small \centering
\begin{tabular}{ll}
(Ref) & je suis \emph{g\^en\'{e}e} que...\\
(BASE) & je suis \emph{embarass\'{e}} que...\\
(TAG) & je suis \emph{embarass\'{e}} que...\\
\end{tabular}}\label{exmpl:not}
\end{li}

For some language pairs the gender-informed system leads to a significant improvement even on a general test set. This implies that the improvement is not merely because of better morphological agreement, as these kinds of improvements are very hard to measure with BLEU, especially given the fact that Europarl consists of formal spoken language and does not contain many sentences using the first person singular pronoun. From our analysis, we observe that in many cases the gender-informed systems have a higher BLEU score than the baseline system due to differences in word choices as in~(\ref{exmpl:wordc}) and~(\ref{exmpl:wordc2}), where both translations are correct, but the gender-informed system picks the preferred variant.

The observations with respect to differences in word preferences between male and female speakers are in accordance with corpus linguistic studies, which have shown that gender does not only have an effect on morphological agreement, but also manifests itself in other ways as males and females have different preferences when it comes to different types of constructions, word choices etc.~\cite{Newman2008,Coates2015}. This also implies that, even for languages that do not mark gender overtly (i.e. grammatically), it can still be beneficial to take the gender of the author/speaker into account.

\begin{li}
{\small \centering
\begin{tabular}{ll}
(Ref) & Je pense que ...\\
(BASE) & Je crois que...\\
(TAG) & Je pense que...\\
\end{tabular}}\label{exmpl:wordc}
\end{li}

Although more research is required in order to draw general conclusions on this matter, from other linguistic studies, it appears that it is indeed the case that there is a relation between the use of the word ``pense'' (``think'') / ``crois'' (``believe'') and the gender of the speaker. To see whether there is a difference in word choice and whether this is reflected in our data, we compiled a list of the most frequent French words for the male data and the female data. Our analysis reveals that ``crois'' is, in general, used more by males (having position 303 in the most frequent words for males, but only position 373 for females), while ``pense'' is found at a similar position in both lists (position 151 and 153). These findings are in accordance with other linguistic corpus studies on language and gender stating that women use less assertive speech~\cite{Newman2008}. ``Croire'' and ``penser'' are both verbs of cognition but there is a difference in the degree of confidence in the truth value predicated: the verb ``croire'' denotes more confidence in the truth of the complement clause than the verb ``penser'' does. In the future, we would like to perform a more detailed analysis of other specific differences in lexical choices between males and females on multiple language pairs.

\begin{li}
{\small \centering
\begin{tabular}{ll}
(Ref) & J' ai plusieurs remarques...\\
(BASE) & J' ai un nombre de commentaires...\\
(TAG) & J' ai plusieurs remarques...\\
\end{tabular}}\label{exmpl:wordc2}
\end{li}

\section{Conclusions and Future Work}\label{sec:conclusions}

In this work, we experimented with the incorporation of speaker-gender tags during the training of NMT systems in order to improve morphological agreement. We focused particularly on language pairs that express grammatical gender but included other language pairs as well, as linguistic studies have shown that the style and syntax of language used by males and females differs \cite{Coates2015}.

From the experiments, we see that informing the NMT system by providing tags indicating the gender of the speaker can indeed lead to significant improvements over state-of-the-art baseline systems, especially for those languages expressing grammatical gender agreement. However, while analyzing the EN--FR translations, we observed that the improvements are not always consistent and that, apart from morphological agreement, the gender-aware NMT system differs from the baseline in terms of word choices. 

In the future, we would like to conduct further manual evaluation on the translations to further analyze the differences with the baseline system. Furthermore, we aim to experiment with other ways of integrating speaker information. We envisage working on gender classification techniques in order to work on other types (more informal) of corpora that are more likely to express speaker characteristics.

\section*{Acknowledgements}\label{sec:ack}

This work has been supported by COST action IS1312, the Dublin City University Faculty of Engineering \& Computing under the Daniel O'Hare Research Scholarship scheme and by the ADAPT Centre for Digital Content Technology, which is funded under the SFI Research  Centres  Programme (Grant  13/RC/2106). Christian Hardmeier was supported by the Swedish Research Council under grant 2017-930.

We would also like to thank the anonymous reviewers for their insightful comments and feedback.

\bibliography{emnlp2018}

\begin{thebibliography}{22}
\expandafter\ifx\csname natexlab\endcsname\relax\def\natexlab#1{#1}\fi

\bibitem[{Bahdanau et~al.(2014)Bahdanau, Cho, and Bengio}]{Bahdanau2014}
Dzmitry Bahdanau, Kyunghyun Cho, and Yoshua Bengio. 2014.
\newblock {Neural Machine Translation by Jointly Learning to Align and
  Translate}.
\newblock In \emph{International Conference on Learning Representations},
  Banff, Canada.

\bibitem[{Cho et~al.(2014)Cho, van Merri{\"{e}}nboer, G{\"{u}}l{\c c}ehre,
  Bahdanau, Bougares, Schwenk, and Bengio}]{Cho2014}
Kyunghyun Cho, Bart van Merri{\"{e}}nboer, {\c C}ağlar G{\"{u}}l{\c c}ehre,
  Dzmitry Bahdanau, Fethi Bougares, Holger Schwenk, and Yoshua Bengio. 2014.
\newblock {Learning Phrase Representations using RNN Encoder--Decoder for
  Statistical Machine Translation}.
\newblock In \emph{Proceedings of EMNLP 2014}, pages 1724–--1734, Doha,
  Qatar.

\bibitem[{Clark et~al.(2011)Clark, Dyer, Lavie, and Smith}]{Clark2011}
Jonathan~H Clark, Chris Dyer, Alon Lavie, and Noah~A Smith. 2011.
\newblock {Better Hypothesis Testing for Statistical Machine Translation:
  Controlling for Optimizer Instability}.
\newblock In \emph{Proceedings of the 49th Annual Meeting of the Association
  for Computational Linguistics: Human Language Technologies: short
  papers-Volume 2}, pages 176--181. Association for Computational Linguistics.

\bibitem[{Coates(2015)}]{Coates2015}
Jennifer Coates. 2015.
\newblock \emph{{Women, Men and Language: A Sociolinguistic Account of Gender
  Differences in Language}}.
\newblock Routledge, London.

\bibitem[{Johnson et~al.(2017)Johnson, Schuster, Le, Krikun, Wu, Chen, Thorat,
  Vi{\'e}gas, Wattenberg, Corrado et~al.}]{GoogleZST}
Melvin Johnson, Mike Schuster, Quoc~V Le, Maxim Krikun, Yonghui Wu, Zhifeng
  Chen, Nikhil Thorat, Fernanda Vi{\'e}gas, Martin Wattenberg, Greg Corrado,
  et~al. 2017.
\newblock {Google's Multilingual Neural Machine Translation System: Enabling
  Zero-Shot Translation}.
\newblock \emph{Transactions of the Association of Computational Linguistics},
  5(1):339--351.

\bibitem[{Klein et~al.(2017)Klein, Kim, Deng, Senellart, and Rush}]{Klein2017}
Guillaume Klein, Yoon Kim, Yuntian Deng, Jean Senellart, and Alexander~M. Rush.
  2017.
\newblock {OpenNMT: Open-Source Toolkit for Neural Machine Translation}.
\newblock In \emph{Proceeding of ACL, Vancouver, Canada}.

\bibitem[{Koehn(2005)}]{Koehn2005}
Philipp Koehn. 2005.
\newblock {Europarl: A Parallel Corpus for Statistical Machine Translation}.
\newblock In \emph{MT Summit}, volume~5, pages 79--86, Phuket, Thailand.

\bibitem[{Litvinova et~al.(2016)Litvinova, Seredin, Litvinova, Zagorovskaya,
  Sboev, Gudovskih, Moloshnikov, and Rybka}]{Litvinova2016}
Tatiana Litvinova, Pavel Seredin, Olga Litvinova, Olga Zagorovskaya, Aleksandr
  Sboev, Dmitry Gudovskih, Ivan Moloshnikov, and Roman Rybka. 2016.
\newblock {Gender Prediction for Authors of Russian Texts Using Regression And
  Classification Techniques.}
\newblock In \emph{Proceedings of the Third Workshop on Concept Discovery in
  Unstructured Data co-located with the 13th International Conference on
  Concept Lattices and Their Applications (CDUD@ CLA)}, pages 44--53, Moscow,
  Russia.

\bibitem[{Mirkin et~al.(2015)Mirkin, Nowson, Brun, and Perez}]{Mirkin2015}
Shachar Mirkin, Scott Nowson, Caroline Brun, and Julien Perez. 2015.
\newblock {Motivating Personality-Aware Machine Translation}.
\newblock In \emph{Proceedings of the 2015 Conference on Empirical Methods in
  Natural Language Processing}, pages 1102--1108, Lisbon, Portugal.

\bibitem[{Newman et~al.(2008)Newman, Groom, Handelman, and
  Pennebaker}]{Newman2008}
Matthew~L Newman, Carla~J Groom, Lori~D Handelman, and James~W Pennebaker.
  2008.
\newblock {Gender Differences in Language Use: An Analysis of 14,000 Text
  Samples}.
\newblock \emph{Discourse Processes}, 45(3):211--236.

\bibitem[{Papineni et~al.(2002)Papineni, Roukos, Ward, and Zhu}]{Papineni2002}
Kishore Papineni, Salim Roukos, Todd Ward, and Wei-Jing Zhu. 2002.
\newblock {BLEU: A Method for Automatic Evaluation of Machine Translation}.
\newblock In \emph{Proceedings of the 40th annual meeting on association for
  computational linguistics}, pages 311--318.

\bibitem[{Pennebaker et~al.(2003)Pennebaker, Mehl, and
  Niederhoffer}]{Pennebaker2003}
James~W. Pennebaker, Matthias~R. Mehl, and Kate~G. Niederhoffer. 2003.
\newblock {Psychological Aspects of Natural Language Use: Our words, Our
  Selves}.
\newblock \emph{Annual review of psychology}, 54(1):547--577.

\bibitem[{Rabinovich et~al.(2017)Rabinovich, Patel, Mirkin, Specia, and
  Wintner}]{Rabinovich2016}
Ella Rabinovich, Raj~Nath Patel, Shachar Mirkin, Lucia Specia, and Shuly
  Wintner. 2017.
\newblock {Personalized Machine Translation: Preserving Original Author
  Traits}.
\newblock In \emph{Proceedings of the 15th Conference of the European Chapter
  of the Association for Computational Linguistics: Volume 1, Long Papers},
  pages 1074--1084, Valencia, Spain.

\bibitem[{Rangel et~al.(2013)Rangel, Rosso, Koppel, Stamatatos, and
  Inches}]{Rangel2013}
Francisco Rangel, Paolo Rosso, Moshe Koppel, Efstathios Stamatatos, and Giacomo
  Inches. 2013.
\newblock {Overview of The Author Profiling Task at PAN 2013}.
\newblock In \emph{CLEF Conference on Multilingual and Multimodal Information
  Access Evaluation}, pages 352--365.

\bibitem[{Santosh et~al.(2013)Santosh, Bansal, Shekhar, and
  Varma}]{santosh2013}
K~Santosh, Romil Bansal, Mihir Shekhar, and Vasudeva Varma. 2013.
\newblock Author profiling: Predicting age and gender from blogs.
\newblock \emph{Notebook for PAN at CLEF}, pages 119--124.

\bibitem[{Sapir(1921)}]{Sapir1921}
Edward Sapir. 1921.
\newblock {Language: An Introduction to the Study of Speech}.
\newblock \emph{NewYork: Harcourt Brace \& Company}.

\bibitem[{Sennrich(2015)}]{Sennrich2015}
Rico Sennrich. 2015.
\newblock {Modelling and Optimizing on Syntactic N-grams for Statistical
  Machine Translation}.
\newblock \emph{Transactions of the Association for Computational Linguistics},
  3:169--182.

\bibitem[{Sennrich et~al.(2016)Sennrich, Haddow, and
  Birch}]{Sennrich2016control}
Rico Sennrich, Barry Haddow, and Alexandra Birch. 2016.
\newblock {Controlling Politeness in Neural Machine Translation via Side
  Constraints}.
\newblock In \emph{Proceedings of the 2016 Conference of the North American
  Chapter of the Association for Computational Linguistics: Human Language
  Technologies}, pages 35--40, Berlin, Germany.

\bibitem[{Slobin(1996)}]{Slobin1996}
Dan~I. Slobin. 1996.
\newblock {From Thought and Language to Thinking for Speaking}.
\newblock In J.~Gumperz and S.~Levinson, editors, \emph{Rethinking Linguistic
  Relativity}, pages 70--96. Cambridge University Press, Cambridge.

\bibitem[{Sutskever et~al.(2014)Sutskever, Vinyals, and Le}]{Sutskever2014}
Ilya Sutskever, Oriol Vinyals, and Quoc~V. Le. 2014.
\newblock {Sequence to Sequence Learning with Neural Networks}.
\newblock In \emph{Advances in Neural Information Processing Systems 27: Annual
  Conference on Neural Information Processing Systems}, pages 3104--3112,
  Montreal, Quebec, Canada.

\bibitem[{Tannen(1991)}]{Tannen1991}
Deborah Tannen. 1991.
\newblock \emph{{You Just Don't Understand}}.
\newblock Ballantine Books, New York, USA.

\bibitem[{Vanmassenhove and Hardmeier(2018)}]{Vanmassenhove2018}
Eva Vanmassenhove and Christian Hardmeier. 2018.
\newblock {Europarl Datasets with Demographic Speaker Information}.
\newblock In \emph{EAMT}, Alicante, Spain.

\end{thebibliography}
\bibliographystyle{acl_natbib_nourl}

\end{document}